# Teach model to answer questions after comprehending the document

Ruiqing Sun[1, 2] and Ping Jian[1, *]

[1] Beijing Institute of Technology, Beijing 100081, China
[2] 2325557558@qq.com
[*] Corresponding author: pjian@bit.edu.cn

**Abstract.** Multi-choice Machine Reading Comprehension (MRC) is a challenging extension of Natural Language Processing (NLP) that requires the ability to comprehend the semantics and logical relationships between entities in a given text. The MRC task has traditionally been viewed as a process of answering questions based on the given text. This single-stage approach has often led the network to concentrate on generating the correct answer, potentially neglecting the comprehension of the text itself. As a result, many prevalent models have faced challenges in performing well on this task when dealing with longer texts. In this paper, we propose a two-stage knowledge distillation method that teaches the model to better comprehend the document by dividing the MRC task into two separate stages. Our experimental results show that the student model, when equipped with our method, achieves significant improvements, demonstrating the effectiveness of our method.

## 1 Introduction

Machine reading comprehension (MRC) is a significant research area within the field of natural language processing (NLP) and holds considerable potential for numerous language-related applications. The primary goal of MRC is to develop systems capable of acquiring knowledge and understanding the underlying logic within documents. In an MRC task, the machine is required to answer one or more questions based on the given documents. To enhance machine capabilities in reading and comprehension, previous researchers have devoted substantial efforts. Early rule-based and traditional machine learning methods improved the machine abilities by focusing on the similarities between candidate answers and the question [1,2,3]. With continued progress in the deep learning field, the context information fusion schemes based on attention mechanisms have led to breakthroughs. Numerous high-performing models have been proposed, such as BiDAF [5] and R-net [6], contributing to the rapid development of the field.

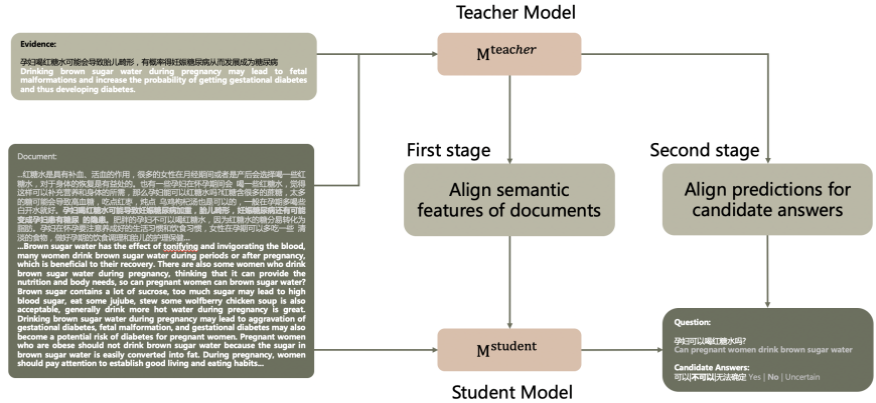

**Fig. 1.** The overview of our two-stage method. Sentences in the picture are adapted from the ReCO dataset [6].

However, these models' unsatisfactory comprehension of language segments still plagues researchers. A new era of pre-training in NLP has been created by models such as GPT [4] and BERT [5]. Unlike previous task driven paradigm, pre-training helps the models better understand natural language text, before they adapt to a specified task. These models have achieved a new significant breakthrough in various NLP tasks, including MRC. Considering the complexity of document expression and the reasoning process that may be required to question answering, the performance of reading comprehension models is still unsatisfactory, along with another challenge: shortcuts. Fortunately, researchers are trying a new direction in MRC that makes the model not only answer the question, but also be able to induce the evidence which can support the answer. By utilizing an MRC corpus annotated with both answers and evidence [6], it is now possible to train MRC models to extract evidence from documents during the answering process. This not only improves the interpretability of answer inference but also addresses the shortcut problem. Typically, in evidence-based MRC, evidence is provided during training but not during testing. Knowledge distillation (KD) is a popular solution, considering the inconsistency between training and testing process [7, 8]. Yu et al. [8] employed distillation on the ReCO [6] dataset to align the student model's features and predictions for candidate answers with those of the teacher model, which proved effective. Specifically, they aimed to encourage the student model to concentrate on words or sentences related to evidence in the document. To accomplish this, they used evidence and candidate answers as input for the teacher model, while documents and candidate answers were used as input for the student model. However, nearly 50% of evidence in the ReCO [6] dataset is manually provided and cannot be directly matched to sentences in the document. Obviously, there is a discrepancy between the information contained in the document and the evidence, which increases the student's learning burden. Moreover, solely using the discriminator to directly align features related to the candidate answers can potentially overlook the semantic information embedded within the document, as evidenced by our experiments detailed in section 4.4. Consequently, their method inevitable faces the problem of taking shortcuts and increases the training complexity. This situation is like a human student learning to solve a math problem by only reading the answer key, rather than consulting the teacher's explanation of the problem-solving approach. The approach mentioned above makes it difficult for the student model to comprehend the document well and make a correct choice from the candidate answers.

In this paper, we introduce a novel approach to enhance the student model's comprehension of the document by dividing the question-answering task into two stages: understanding the text and providing the answer. Specifically, in the first stage, we align the features of the student and teacher models that represent the semantics embedded in the document. In the second stage, we align the predictions for the candidate answers, as illustrated in Figure 1. The purpose of this approach is to give equal weight to the model's ability to comprehend semantic information and answer questions. Moreover, unlike previous methods, we add documents to the input of the teacher model to minimize the information gap between the inputs of the teacher and student models. This approach not only improves the teacher model's performance by providing comprehensive semantic and contextual information, but also benefits the student model, as high-quality teachers tend to produce better apprentices. The division of this approach is inspired by two-stage object detection methods commonly used in the field of computer vision (CV). These methods aim to accurately identify and localize objects within images through a two-step process, which are relatively independent. The first stage generates a set of candidate regions or proposals likely to contain objects, while the second stage refines these proposals and classifies them into specific object categories. In some cases, researchers even pre-train the first part of the model on unrelated datasets before fine-tuning on specific tasks. For instance, P2Pnet [9], proposed to address the challenge of crowd counting, trained its feature extraction component on the ImageNet [10] dataset before fine-tuning.

Our work can be summarized in the following three aspects:

1. In this paper, we reviewed the multi-choice MRC task from the perspective of object detection tasks and proposed that the multi-choice MRC task should be divided into two stages: semantic comprehension and question answering.

2. We propose a two-stage distillation method that pays equal attention to the student model's ability to comprehend semantics and answer questions correctly. By employing our method, the student model achieves state-of-the-art (SOTA) results.

3. To create an intelligent teacher and reduce information gaps between the two models, we combine evidence and documents as the input for the teacher model. Our experimental results show that teacher models with inputs consisting of both evidence and documents outperform those from previous methods.

## 2    Related Work

### 2.1    Machine Reading Comprehension

The MRC tasks can be roughly divided into several categories, such as multi-choice [11, 12], span prediction [13, 14, 6], free form [15], knowledge-based MRC [16], multi-passage MRC [18-1], MRC with unanswerable questions [18, 19] and so on. To address the challenges of these tasks, early rule-based and machine learning methods attempted to provide answers based on the semantic similarities between candidate answers and questions. Although these approaches contributed to the development of research in related fields, they suffered from limited robustness and low accuracy. The release of large-scale datasets like Daily Mail [12], SQuAD [20], and MS MARCO [21] made it possible to tackle MRC tasks using deep neural network frameworks. Some methods attempt to model human reading modes by initially extracting evidence from documents [22, 23, 24, 25, 26]. It can be argued that extracting evidence before making a choice could improve model performance and enhance interpretability. Evidence extraction methods in MRC can be broadly classified into supervised, unsupervised, and other methods, based on differences during training phases. Supervised methods typically require training models with expert-provided or automatically-generated gold-standard evidence, while unsupervised methods do not provide gold-standard evidence. However, obtaining gold-standard evidence can be time-consuming and labor-intensive, and the performance of an MRC system comprising two models with independent targets can be difficult to evaluate simultaneously. These limitations hinder the advancement of related methods and diminish their practical applications. Therefore, alternative approaches have been proposed, such as weakly supervised methods that rely on noisy evidence labels obtained through heuristic rules [25].

## 2.2 Knowledge Distillation

Knowledge distillation (KD) [27] is an effective method that enables student models to learn knowledge or specific capabilities from teacher models. Distillation using soft label knowledge is a simple and effective KD method which is easily combined with other distillation methods. It does not require consideration of specific structural differences between teacher and student models. However, soft labels often provide limited information and contain noise which may enhance the learning difficulties. DFA [28] is a method designed to address this issue by distilling intermediate layer knowledge from the teacher model. It improves the performance by enabling the student model to learn feature expression capabilities from the teacher model's intermediate hidden layers. What's more TinyBERT [30] and MobileBERT [31] reduce the number of parameters in BERT by adjusting the depth and width dimensions of hidden layers through knowledge distillation. These lightweight models can be better applied in resource-constrained scenarios and achieve performance on par with their teacher. In the past, it was widely believed that human students needed far fewer training examples than learning machines. To address this challenge, researchers proposed the Learning Using Privileged Information (LUPI) [29] method. Unlike the previous approaches that approximate the unknown decision rule as closely as possible, LUPI makes the teacher more intelligent and provides additional information. To enhance the student model's performance on the ReCO [6] dataset, Yu et al. [8] aligned the output representations of the special token [OPT], in addition to aligning the soft labels. They provided the teacher with additional evidence as input, which is invisible to the student.

Unlike the previous work, we divide the MRC task into two stages: semantic comprehension and question answering. The student model is encouraged to focus on understanding the document rather than producing a specific evidence output.

## 3 Method

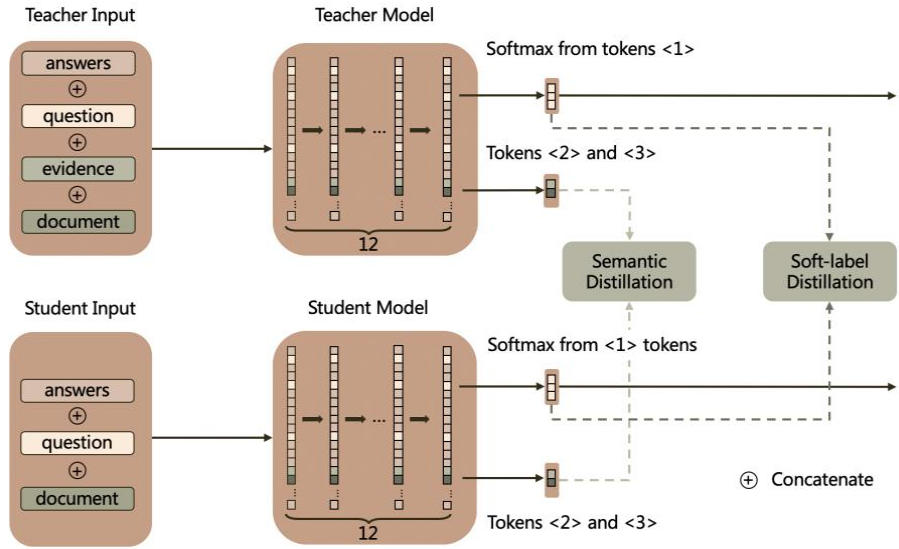

**Fig. 2.** The two distillation stages of our method. In the first stage, we perform distillation using the output semantic representations from tokens <2> and <3> while using the output softmax generated from tokens <1> in the second stage.

Although many previous models, such as BERT [5], consist of hidden layers (12 in BERT-base) that have nearly the same architecture, we propose that the multi-choice MRC task should be considered as two relatively independent stages. Specifically, the student model should first comprehend the semantics contained in the document based on the question and then select the correct answer from the candidates.

### 3.1 Teacher Model

A good teacher model is an essential component of a successful distillation process. As a result, we constructed a teacher model with input derived from concatenating the candidate answers, question, evidence, and document. In other words, we provide the teacher model with contextual and background information that can help improve understanding of evidence, which is quite distinct from Yu et al.'s previous work [8]. In order to prove that the performance improvement of the model is derived from our proposed method rather than other factors, we followed earlier approaches and trained the teacher model by minimizing the cross-entropy loss function, as demonstrated in Equation (1).

$$L_{CE}(y, p) = -\sum_{i=0}^{C-1} y_i \log(p_i) \quad (1)$$

Where y is a one-hot vector representing the sample label, like $[y_0 ... y_{c-1}]$. $p_i$ represents the probability that the sample belongs to the i-th category, and C is the number of categories.

### 3.2 Student Model

Unlike the teacher model, the input for the student model consists of candidate answers, questions, and the document. In other words, our goal is for the student model to ultimately develop the ability to comprehend the input document and select the correct answers without requiring additional evidence. To simplify the knowledge distillation processes and enhance the practical value of our method, we choose BERT-base as the backbone for both models. The sole distinction between them lies in whether or not the input includes evidence, as shown in Fig 2.

### 3.3 The First Stage: Semantic Comprehension Distillation

Using specific tokens to represent sentence features is a common approach for addressing downstream MRC tasks. To enhance the student model's ability to understand documents, we align the output representations of the token <2> as well as the token <3>. The token <2> is used to concatenate the candidate answers with other sentences, while the token <3> is used to indicate the start of question sentences. By minimizing the MSE loss of these output representations, the student model distills the teacher's knowledge to learn the semantic comprehension ability, as illustrated in Fig. 2. The loss function for this stage can be expressed as (2).

$$L_{MSE} = \frac{1}{N}\sum_{i=1}^{N}(T_i - S_i) \quad (2)$$

where N is the number of samples. T and S represent the output representations of the teacher and student model respectively.

### 3.4 The Second Stage: Soft Label Distillation

In this stage, we expect the student model to learn the ability to choose the correct answer from the teacher after comprehending the documents. To achieve this, we train the student model by additionally minimizing the KL divergence of the output softmax between the two models that are generated from tokens <1>, as illustrated in Fig 2. The tokens <1> are used to concatenate the candidate answers. The loss function at this stage can be expressed by equation (4), where T represents the output representations of the teacher model and S represents those of the student model. The α and β are two hyper parameters that can be adjusted. After completing the two stages of KD, we obtain the final student model with both semantic comprehension and candidate selection abilities.

$$L_{KL}(T /\!/ S) = \sum_{i=1}^{N} T(x_i) \cdot (\log T(x_i) - \log S(x_i)) \quad (3)$$

$$L_{SLD}(T /\!/ S, y, p) = \alpha L_{KL}(T /\!/ S) + \beta L_{CE}(y, p) \quad (4)$$

## 4 Experiments

In this paper, we follow previous work and use accuracy as the evaluation metric. All experiments were conducted on an RTX 3090(24G) with a batch size of 16, and optimized using AdamW provided by PyTorch. The weight_decay parameter value was set to 0.01, while the default values for other parameters were maintained. Our model was initialized using the pre-trained model from HuggingFace's Transformer library.

### 4.1 Datasets

We carried out extensive experiments on the ReCO [6] dataset to demonstrate the effectiveness and validity of our proposed method from various perspectives. The ReCO dataset was introduced to address the issue of the comprehension process being reduced to a retrieval process in many previous datasets due to irrelevant information in the questions [6]. Questions in ReCO are opinion-based real-world queries that can be either factoid or non-factoid and can be answered with yes/no/uncertain. This makes the reasoning process of machine learning methods built upon this dataset more credible than many others. The ReCO [6] dataset comprises 280k training samples and 20k test samples, which are further split into Test Set A (TestA) and Test Set B (TestB). We used TestA in this study, following previous works. It is worth mentioning that during the annotation process, annotators paraphrased or highly summarized key sentences for 46% of the samples based on their understanding. This demands that the student model enhance its comprehension of the document, as the evidence cannot be directly located within it.

**Table 1.** Performance of some prevalent and we deployed models.

| Model | Performance on Testa (only evidence) | Performance on Testa (only document) |
|---|---|---|
| Random | 33.3 | 33.3 |
| BiDAF | 68.3 | 55.8 |
| BiDAF* | 70.9 | 58.9 |
| BERT-base | 73.4 | 61.1 |
| BERT-large | 77.0 | 65.3 |
| ALBERT-tiny | 70.4 | 62.7 |
| ALBERT-base | 77.6 | 68.4 |
| MSKDTS | 77.6 | 71.0 |
| BERT-base (our deployment) | 76.7 | 70.2 |
| human | 91.5 | 88.0 |

### 4.2 Baseline

To evaluate the effectiveness of our method, we selected several strong baselines that perform well on various MRC tasks. The metrics of these prevalent models are obtained from previous research [8].

BiDAF [23]: BiDAF employs LSTM as its encoder and models the relationship between the question and the answer using a bidirectional attention mechanism.

BiDAF* [23, 32]: BiDAF* replaces the traditional word embedding in BiDAF with ELMo (a language model trained on unsupervised data), yielding improved results.

BERT [5]: A multi-layer bidirectional Transformer, BERT is pre-trained on large unlabeled data and has outperformed state-of-the-art models in numerous NLP tasks.

ALBERT [33]: ALBERT is an enhanced version of BERT, reducing the overall number of parameters, accelerating the training process, and surpassing BERT in many aspects.

Table 2. Performance of teacher model.

| Model | Performance on Testa |
| --- | --- |
| Random | 33.3 |
| BERT (only document) | 76.7 |
| Teacher (with evidence guidance) | **78.2** |
| human | 91.5 |

Table 3. The accuracy of the student trained by different methods.

| Model | Performance on Testa (only document) |
| --- | --- |
| BERT-base (without teacher) | 70.2 |
| Student model (with randomly linear layer) | 70.8 |
| Student model (without linear layer) | 44.6 |
| Student model (LMSKDTS) | 71.0 |
| Student model (single-stage) | 71.3 |
| Student model (distillation*) | 70.8 |
| Student model (two-stage) | **71.8** |
| human | 88.0 |

To more effectively demonstrate the effectiveness of our method for both teacher and student models, we also deployed BERT on our machine. Since the original BERT was not designed to address multi-choice tasks, we added a linear layer without activation functions to make it more suitable. All performances are shown in Table 1. The second/third column displays the results of these models when taking evidence/documents as input for both the training and testing phases.

### 4.3 Teacher Model

Unlike previous work, we included the document in the input for getting an intelligent teacher. The output softmax genenate from tokens <1> were used to select the correct answer. The experiment shows that using documents to assist the teacher effectively improves their performance on the ReCO dataset, as seen in Table 2. The initial learning rate for this experiment was set to 8e-5, with a total of 10 epochs of training, taking approximately 500 minutes. The learning rate was reduced to half of its original value in the third epoch and further reduced to a quarter in the sixth epoch. To ensure the credibility of our work, all training settings are consistent for both BERT we deployed and the teacher model. It is worth mentioning that we saved the trained model after the 10th epoch as the teacher for later processes, rather than the best one, to avoid test sample leakage.

### 4.4 Two-stage Distillation

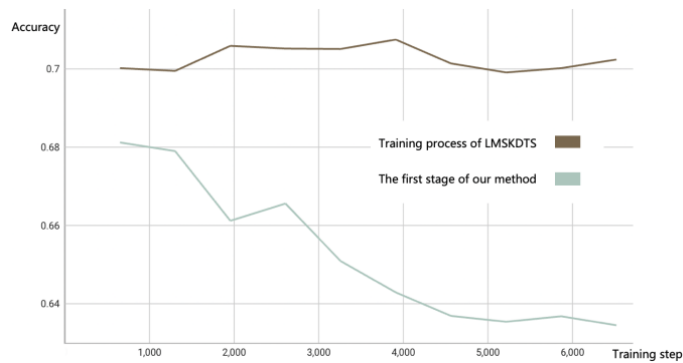

**Fig. 3.** The performances of both the student model and LMSKDTS during the training processes.

We propose dividing the MRC task into two relatively independent stages: semantic comprehension and question answering. Through our two-stages distillation method, the student model's performance improves effectively. In this set of experiments, we aim to demonstrate that the improvement in the student model comes from better semantic comprehension. To achieve this, we use a method similar to MSKDTS [8], which is named LMSKDTS by us, to align the output representations of token <1> in the first stage and train through the soft label distillation method in the second stage. Token <1> is used to concatenate candidate answers, which is also the same as in MSKDTS. As the proposed LMSKDTS method has nearly the same architecture and accuracy as MSKDTS, it can be considered a reproduction of MSKDTS. Therefore, we use it as a comparison to show the advantages of our work.

Interestingly, even when using a linear layer with randomly initialized parameters to choose the answer after aligning the output representations of token <1>, the model is still capable of answering most questions. Specifically, the student model trained by LMSKDTS with an untrained linear layer performs on par with MSKDTS [8] with nearly 70.8% accuracy. However, unlike the above-mentioned case, the performance of the student trained by our proposed method gradually decreases during the first distillation stage's training process in our two-distillation method, as shown in Fig. 3. To investigate this further, we modified the original layer to directly sum the model's output representations of token <1>. The model still has a preliminary ability to answer questions in this situation, as shown in Table 3. This strongly confirms that aligning the representations of token <1> is likely more focused on how to choose the correct answer rather than comprehending the document's semantics.

### 4.5 Compared with Single-stage Distillation

In this section, we demonstrated that the improvement in the student model is primarily attributed to the two-stage distillation process rather than the alteration of the loss function. We constructed a single-stage loss function by combining all the losses from the two-stages method, as illustrated in equation (5), and implemented it in a single-stage method.

$$L_{total}=\alpha L_{CE}+\beta L_{KL}+\gamma L_{MSE} \qquad (5)$$

where α, β and γ are tunable parameters to balance the three targets of training phase. We set α to 0.25, β to 0.25 and γ to 0.5.

The single-stage method involves only one training stage and aligns the representations of tokens related to semantics and predictions simultaneously, as opposed to sequentially in our two-stage method. We trained the model using the single-stage method for over 60 hours, which is slightly longer than the two-stage methods. However, the performance of this method was underwhelming, as shown in Table 3. This result can be attributed to the high complexity of the loss function and the fact that the soft labels and semantic features provided by the teacher model inevitably contain some noise. This increased learning difficulty and led to confusion for the student model during the training process. In other words, learning in two stages is much easier for the student.

### 4.6 Align the Features from Different Hidden Layers

In this section, we demonstrated the student model has learned the ability to comprehend the document and it is meaningless to blindly aligning more output from hidden layers of two models. During the experiment, we tried to align the output representations of tokens <2> and <3> from both the 7th and last layers in the first stage to further emulate the teacher model's behavior. We referred to this method as distillation*. The performance of distillation* even becomes worse when compared with the original two-stage distillation method, as shown in Table 2. This highlights that blindly imitating the teacher model's behavior is not effective and the success of our two-stage distillation method can be attributed to the student model's enhanced ability to comprehend semantics, as higher layers effectively capture semantic features [34].

## 5 Conclusion and Future Work

In this paper, we reviewed the multiple-choice MRC task by drawing inspiration from the computer vision field and proposed a two-stage knowledge distillation method that achived SOTA performance. Our method encourages the student to learn the ability to comprehend the document's semantics in the first stage and the ability to select the correct answer from the candidates in the second stage. The effectiveness of our method has been validated through various experiments on the ReCO dataset. We believe our method can be easily extended to other MRC tasks or to train other model beyond BERT, as its training procedure is easy to implement. In future work, we aim to further investigate what the student model has genuinely learned from the teacher model in order to enhance the interpretability of our method.